\documentclass[letterpaper, 10 pt, conference]{ieeeconf}  
\usepackage{amsmath}

\IEEEoverridecommandlockouts                              

\usepackage{graphics} 
\usepackage{graphicx}
\usepackage{cite}
\usepackage{comment}

\title{\LARGE \bf
Robotic Fruits with Tunable Stiffness and Sensing: Towards a Methodology for Developing Realistic Physical Twins of Fruits}

\author{Saitarun Nadipineni$^{1}$, Keshav Pandiyan$^{1}$, Kaspar Althoefer$^{1}$, Shinichi Hirai$^{2}$, Thilina Dulantha Lalitharatne$^{1}$
\thanks{*This work was supported by Royal Society UK International Exchange Grant (IES$\backslash$R1$\backslash$251574) and partially supported by the PALPABLE project, funded by the UK Research and Innovation (UKRI) through the UK government's Horizon Europe funding guarantee (grant No.101092518) and funded by the European Union.}
\thanks{$^{1}$School of Engineering and Materials Science,
       Queen Mary University of London, Mile End Rd, Bethnal Green, London E1 4NS, United Kingdom.
        {\tt\small \{s.nadipineni, ex24378, k.althoefer, t.lalitharatne\}@qmul.ac.uk}}%
\thanks{$^{2}$Department of Robotics, Ritsumeikan University, Japan
        {\tt\small hirai@se.ritsumei.ac.jp}}%
}

\begin{document}
\maketitle
\thispagestyle{empty}
\pagestyle{empty}

\begin{abstract}

The global agri-food sector faces increasing challenges from labour shortages, high consumer demand, and supply-chain disruptions, resulting in substantial losses of unharvested produce. Robotic harvesting has emerged as a promising alternative; however, evaluating and training soft grippers for delicate fruits remains difficult due to the highly variable mechanical properties of natural produce. This makes it difficult to establish reliable benchmarks or data-driven control strategies. Existing testing practices rely on large quantities of real fruit to capture this variability, leading to inefficiency, higher costs, and waste. The methodology presented in this work aims to address these limitations by developing tunable soft physical twins that emulate the stiffness characteristics of real fruits at different ripeness levels. A fiber-reinforced pneumatic physical twin of a kiwi fruit was designed and fabricated to replicate the stiffness at different ripeness levels. Experimental results show that the stiffness of the physical twin can be tuned accurately over multiple trials (97.35 - 99.43\% accuracy). Gripping tasks with a commercial robotic gripper showed that sensor feedback from the physical twin can reflect the applied gripping forces. Finally, a stress test was performed over 50 cycles showed reliable maintenance of desired stiffness (0.56 - 1.10\% error). This work shows promise that robotic physical twins could adjust their stiffness to resemble that of real fruits. This can provide a sustainable, controllable platform for benchmarking and training robotic grippers.

\end{abstract}

\section{INTRODUCTION}

In recent years, the global agri-food industry has been under severe pressure due to factors such as labour shortages, increasing consumer demands, and pandemics \cite{win2025labor,saitone2017agri,luckstead2021labor}. The inefficiency of manual food harvesting is evident here, resulting in wasted resources, effort, and economic loss. In recent years, robotic harvesting has proven to be a promising solution to address the limitations of manual harvesting. As a result, a variety of rigid and soft robotic grippers were developed to harvest soft fruits and vegetables \cite{gunderman2021tendon,navas2021soft,johnson2024field}. In addition, these grippers were paired with mobile robotic manipulators and mobile robots to perform fruit harvesting in fields \cite{junge2023lab2field}. However, challenges in testing and evaluating novel grippers limit their development \cite{kootstra2021selective}. 

 Robotic manipulation in the fruit harvesting and food industry is inherently complex due to the fragile, deformable, and highly variable nature of biological materials. The stiffness of a fruit is influenced by multiple factors, including temperature, humidity, and ripeness, since environmental conditions have been shown to alter viscoelastic responses of fruit tissues \cite{matas2005humidity}. Food products differ widely in geometry, texture, moisture, and surface condition, making it difficult for robotic systems to achieve human-level adaptability in grasping and recognition tasks \cite{wang2022challenges}. The development and validation of such systems often require large quantities of real fruits and vegetables for repeated testing, resulting in substantial economic and environmental waste. Furthermore, each trial is affected by uncontrollable factors, such as temperature, humidity, and ripeness, leading to inconsistent mechanical responses \cite{Qiu2023evaluation}. These variations hinder effective benchmarking and the implementation of data-driven or control-based training strategies for novel grippers \cite{zhou2022intelligent}. Combined with the high cost and complexity of integrating vision, sensing, and soft end-effector technologies, this makes robotic manipulation of food one of the most challenging areas of automation. Consequently, ensuring reproducible and sustainable benchmarking across experiments requires exploring alternative testing platforms that can replicate fruit mechanics under controllable and repeatable conditions.

\begin{figure*}[h!]
    \centering
    \includegraphics[scale=0.57]{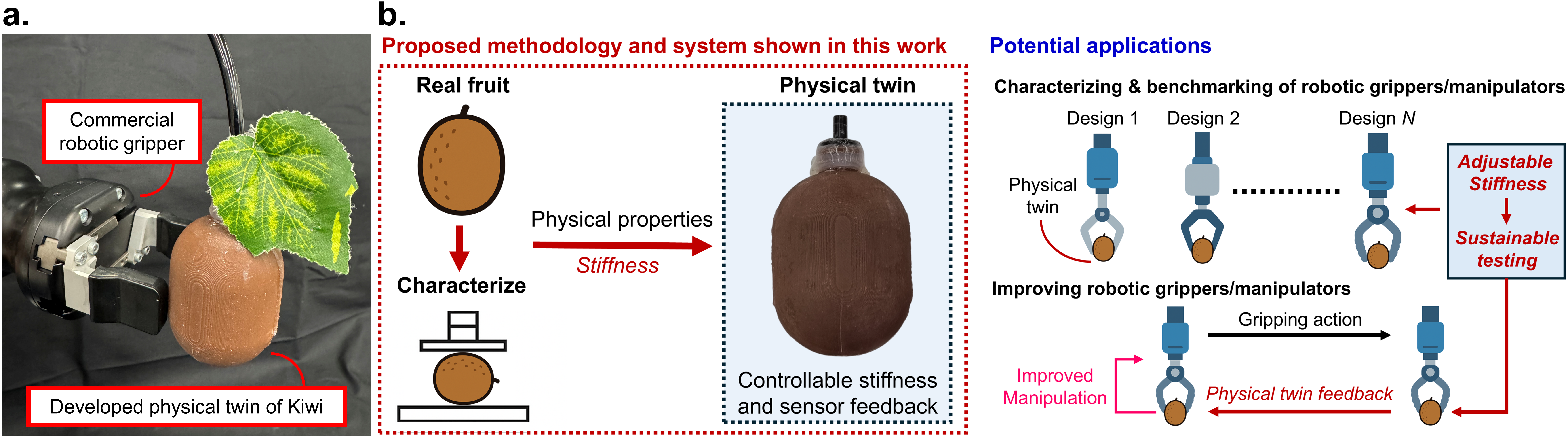}
    \caption{Overview of the proposed methodology and potential applications. (a) The developed soft physical twin of a kiwi fruit is being grasped by a commercial robotic gripper. (b) The proposed methodology begins with the mechanical characterization of real fruits to extract stiffness properties, which can be replicated through a robotic physical twin with controllable stiffness and integrated sensor feedback. The developed twin has potential applications for sustainable, repeatable testing to characterize, benchmark, and improve various robotic grippers/manipulators through adjustable stiffness and sensor feedback. \textit{The illustration of the Kiwi fruit in (b) and the grippers in (c) was generated using ChatGPT.}}
    \label{Fig_1}
\end{figure*}

Physical twins can be an attractive solution to this problem, as they mimic the physical properties of real-world objects. Such systems have also been implemented in the medical field to replicate patient-specific physiological and behavioural responses. For example, a recent study \cite{he2021hapticmouse} developed a tunable-stiffness haptic palpation simulator to reproduce muscle guarding during abdominal examination, while another \cite{lalitharatne2021morphface} introduced MorphFace, a hybrid morphable robotic face capable of rendering patient pain expressions across different genders and ethnicities. These studies demonstrate the potential of controllable physical interfaces to emulate biological tissue and perceptual feedback for clinical training. Similarly, sensorized physical twins of foods can enable effective, sustainable training of robotic grippers by replicating food properties and providing feedback. More recently, this approach expanded into the agricultural domain to develop a soft physical twin of a raspberry, leveraging fluidic sensing \cite{junge2022soft}. This prior work aimed to develop a physical twin that can measure the pulling forces when the twin is attached and detached from the plant. This aided the development of an effective control policy for harvesting \cite{junge2023lab2field}. 

However, to the best of our knowledge, physical twins that can mimic the properties of real fruits are absent in the literature. This is an important problem to tackle, as physical twins capable of tuning their stiffness can provide effective training data to use gripping forces appropriate to the ripeness level of fruits. For instance, when a fruit is ready to be harvested, it is very raw, hence higher gripping forces can be used to pluck the fruit without slippage. On the other hand, when fruits are packaged and sorted for shipment to stores, they are riper; hence, smaller gripping forces are required. Physical twins with elastic properties can support sustainable training, benchmarking of novel grippers, and the development of control policies for food manipulation. They can recover from faulty gripping forces and handling without imposing the wastage of effort, cost, and time like real fruits.

To this end, we propose a methodology for developing tunable physical twins of fruits that simulate different levels of ripeness based on actual fruits (Fig. \ref{Fig_1}(a-b)). In this work, we developed a fiber-reinforced soft physical twin of a Kiwi fruit that can adjust its stiffness to mimic different levels of ripeness. In the remainder of the paper, we first present the stiffness characterization of Kiwi fruits, which helped guide the design of the physical twin and provide insight into their stiffness profiles. Then, we present the design and fabrication process of the physical twin in section \ref{Design_fabrication_sec}. In section \ref{experiments_sec}, we present the method for tuning and evaluating the stiffness of the physical twin. We also performed a simple gripping task with a commercial gripper to show the sensory feedback that can be achieved through the physical twin based on the gripping forces. In the final experiment, we performed a stress test to evaluate the repeatability and consistency in maintaining desired stiffness during multiple loading and unloading cycles. The findings are also discussed in this section. Finally, in section \ref{limitations_future_work_sec}, we discuss the limitations and future directions of this work and conclude in section \ref{Conclusion_sec}.  

\begin{figure*}[h!]
    \centering
    \includegraphics[scale=0.5]{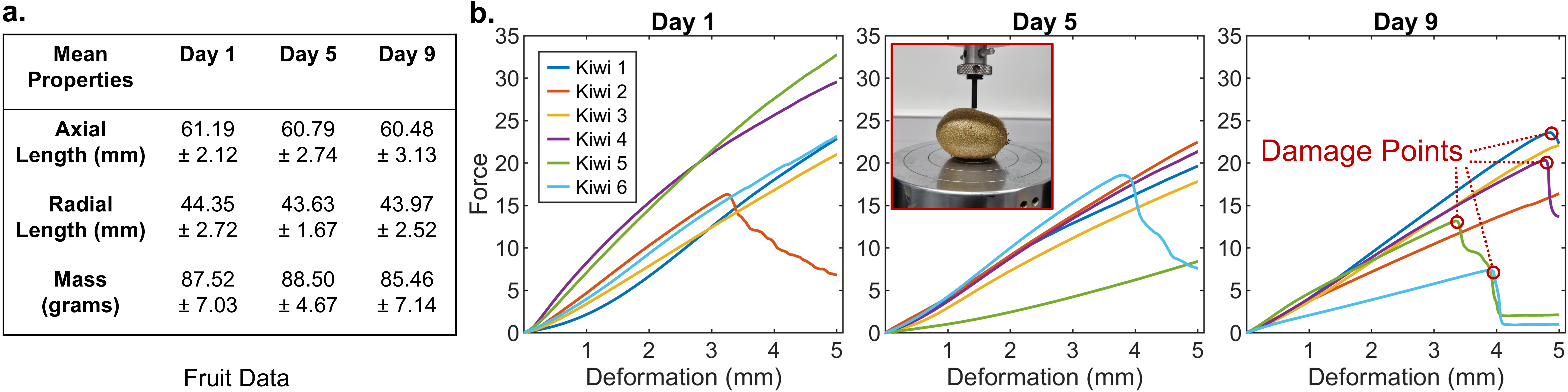}
    \caption{Mechanical characterization of real kiwi fruits across ripening stages. (a) Mean geometric and mass properties of six kiwi samples measured on Day 1 (raw), Day 5 (mid-ripe), and Day 9 (ripe). (b) The force–deformation responses from indentation tests on each day are shown here. Each curve represents one fruit sample indented to 5 mm at 5 mm/min. The experimental setup is also presented here, where a cylindrical indenter is mounted to an Instron 68-TM testing machine. Progressive softening with ripening is evident from the reduction in slope. Damage points on Day 9 indicate early failure in over-ripe fruits compared to other testing days.}
    \label{Ground_truth_fig}
\end{figure*}

\section{METHODS}

\subsection{Evaluating the stiffness of Kiwi fruits} \label{real_fruits_characterization}
The kiwifruit was chosen for this study due to its near-ellipsoidal geometry, which provides a simple and well-defined shape for geometric modelling and silicone-based fabrication. To develop the physical twin, a single mechanical profile encompassing the entire fruit is required. Prior work has examined its mechanical behavior through various methods. Literature states that fruit hardness was related to tissue mechanics when tensile tests were performed on peel strips of Kiwi fruits and compression tests were performed on their flesh cubes \cite{huang2022hardness}. The findings existing literature state that peel moduli was $\approx7.5$--$21\mathrm{MPa}$, tensile strengths were $\approx0.5$--$1.1\mathrm{MPa}$, and the reduction in flesh modulus was from $\approx1.15$ to $0.26\mathrm{MPa}$ with ripening ($R^{2}=0.97$--$0.99$) \cite{huang2022hardness}. Another study noted the whole-fruit failure loads to be $\approx180$--$250\mathrm{N}$ \cite{URBANSKA2025113682}. Tactile sensing with Hertz theory was used in another study to estimate stiffness ($\approx0.4$--$1.5\mathrm{N,mm^{-1}}$, $\leq2\%$ error) \cite{Erukainure9863245}. Finite-element analysis was also performed in existing literature to determine the bruise thresholds \cite{Zhu13050785}.

While these studies provide valuable insight, major limitations persist. The reported mechanical properties vary widely between experiments due to differences in cultivar, ripeness, hydration, and temperature. Most adopt linear-elastic assumptions, neglecting viscoelastic and anisotropic effects that dominate biological tissues. Flat-plate and indentation setups are highly sensitive to alignment and surface curvature, producing significant variability in force readings. Coupon-level tests separate peel and flesh, destroying their native constraints and mechanical coupling, whereas finite-element methods depend heavily on assumed material laws and calibration accuracy. Such inconsistencies hinder reproducibility and make it difficult to establish unified stiffness benchmarks. Consequently, we focus on direct comparison of indentation forces at a specific deformation as a practical indicator of stiffness, minimizing analytical assumptions while capturing the fruit’s natural response relevant to robotic manipulation.

The kiwi fruits characterized in this study were of the Hayward variety, commercially sourced from a local supermarket. We purchased three packs, each containing 6 fruits. The fruits were all unpacked and mixed to allow for a fair, generalizable test, reducing biases introduced by a specific batch. We tested 6 random fruits on three different days. We chose to maintain a 3-day gap between testing days to allow the fruits to ripen significantly. Therefore, the experimental results can be categorized into Day 1 (day of purchase), Day 5, and Day 9. We recorded the axial length, radial length, and mass of each fruit on the day of testing (reported in Fig. \ref{Ground_truth_fig}(a)). Then, we performed indentation tests using a 5 mm-radius cylindrical indenter and an Instron 68-TM materials testing machine with a 2 kN force sensor. Each fruit was deformed by 5mm at an indentation rate of 5mm/min, with a preload of 0.1 N applied before each test. Fig. \ref{Ground_truth_fig}(b) shows the results of the indentation tests on each day. Between testing days, the fruits were stored in a closed container at room temperature (23-24 \textdegree C).  

\begin{table}[h!]
    \centering
    \normalsize
\caption{Stiffness of Kiwi fruits at 3 mm Deformation}
\label{Kiwi_data_table}
    \begin{tabular}{ccc}
         \hline 
         \textbf{Days} &  \textbf{Mean Stiffness (N/mm)} &  \textbf{STD (N/mm)}\\
         \hline\hline 
         1  &  5.41 & $\pm$ 1.36\\
         5  &  3.94 & $\pm$ 1.32\\
         9  &  3.89 & $\pm$ 1.07\\
         \hline
    \end{tabular}
\end{table}

From the results shown in Fig. \ref{Ground_truth_fig}(b), it is clear that the force-deformation relationship exhibits strong linearity. Hence, it can be assumed that the Kiwi fruits deform linearly; therefore, their stiffness can be found through $K = F_{R} / d$, where $K$ is the stiffness, $F_{R}$ is the reaction force, and $d$ is the deformation. However, the results also show that some Kiwis reach their damage points before they were indented by 5 mm. From Days 1 and 5, only one Kiwi reached its damage point. Whereas on Day 9, four Kiwis reached their damage point. This shows that ripper Kiwis reach their damage points earlier than raw Kiwis. With this limitation in mind, we chose to use the force values at 3 mm deformation to find the stiffness. This is because none of the Kiwis reached their damage point before 3 mm. Under the linear-deformation assumption, the stiffness is constant until the Kiwis reach their damage point. The mean stiffness values are presented in Table \ref{Kiwi_data_table}, along with the standard deviations for each testing day. These stiffness values can be used to tune the physical twin to simulate each ripeness stage.   

\subsection{Design and fabrication of physical twin}\label{Design_fabrication_sec}

\begin{figure}[h!]
    \centering
    \includegraphics[scale=0.46]{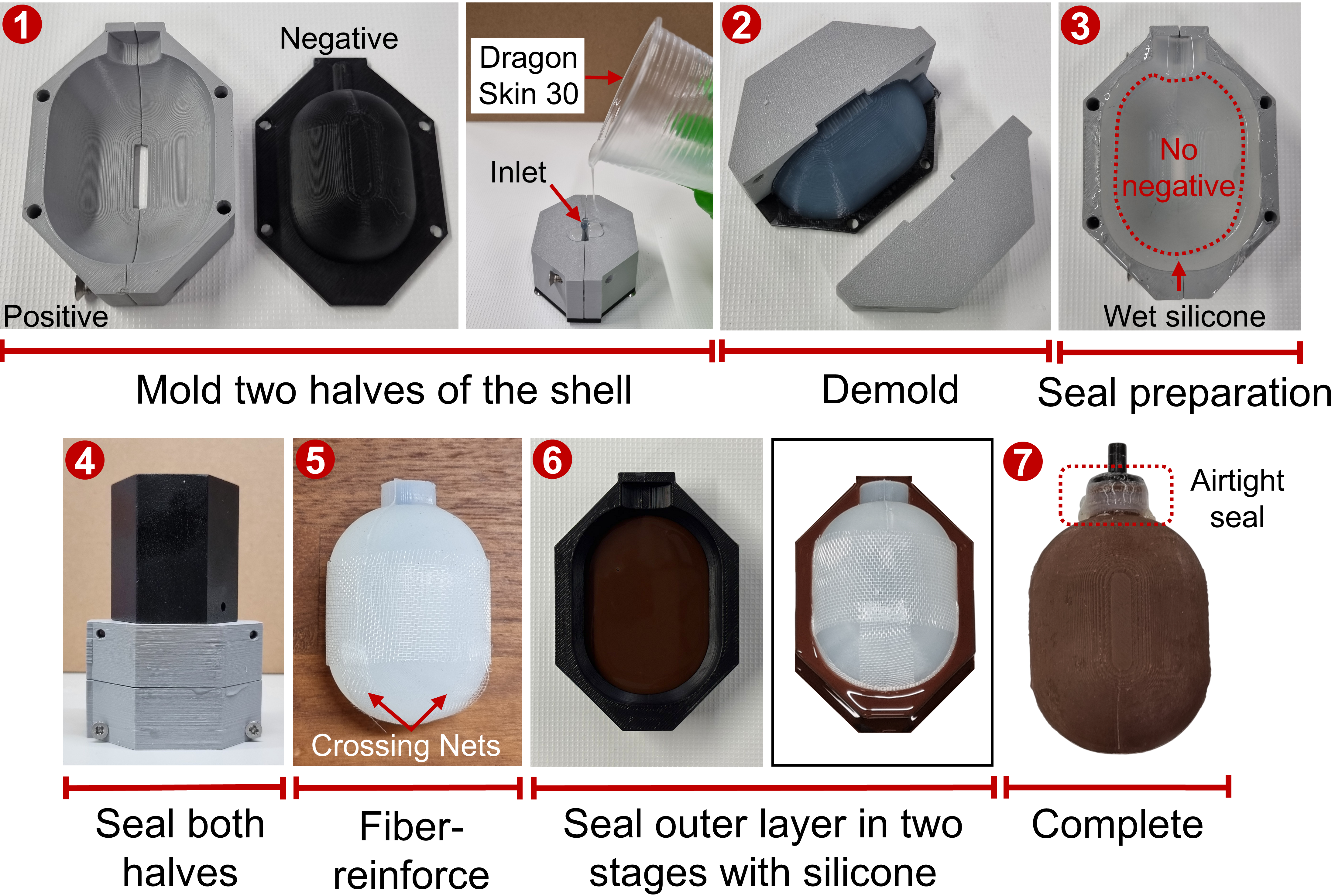}
    \caption{Fabrication process of the soft kiwi physical twin. Step-by-step fabrication showing (1) mold preparation for casting two silicone pieces, which primarily make up the main body of the physical twin, (2) demolding, (3) Seal preparation to join the two silicone pieces placed in their molds, (4) sealing of both halves securely by placing a mass on top, (5) fiber-reinforcement using crossing nets, and (6,7) final outer-layer sealing to form the complete fruit.}
    \label{Fabrication_fig}
\end{figure}

The physical twin prototype was fabricated with Dragon Skin 30 (SmoothOn, Inc.) silicone rubber. First, a two-part separable mold was used to produce two identical silicone pieces. These pieces act as the main body of the physical twin. Once the silicone pieces were cured, the two halves were demolded, and the negative part was removed. Then, each piece was placed into its mold again, and joined using Dragon Skin 30 as an adhesive. A mass was placed on top of the two molds, which were pressed together to ensure the pieces joined properly. Once the seal cured, three pieces of garden net were used to fiber-reinforce the silicone body. The ends of each piece were attached to the silicone body with small drops of super glue. Finally, the top and bottom sides of the silicone body were placed in a mold with Dragon Skin 30 mixed with Silc-Pig Brown (SmoothOn, Inc.) and were cured. Finally, cable ties, Silpoxy (SmoothOn, Inc.), and a rapid two-part epoxy were used to minimize air leakage at the 3D-printed air inlet. The manufacturing steps are illustrated in Fig. \ref{Fabrication_fig}.

\begin{figure*}[h!]
    \centering
    \includegraphics[scale=0.55]{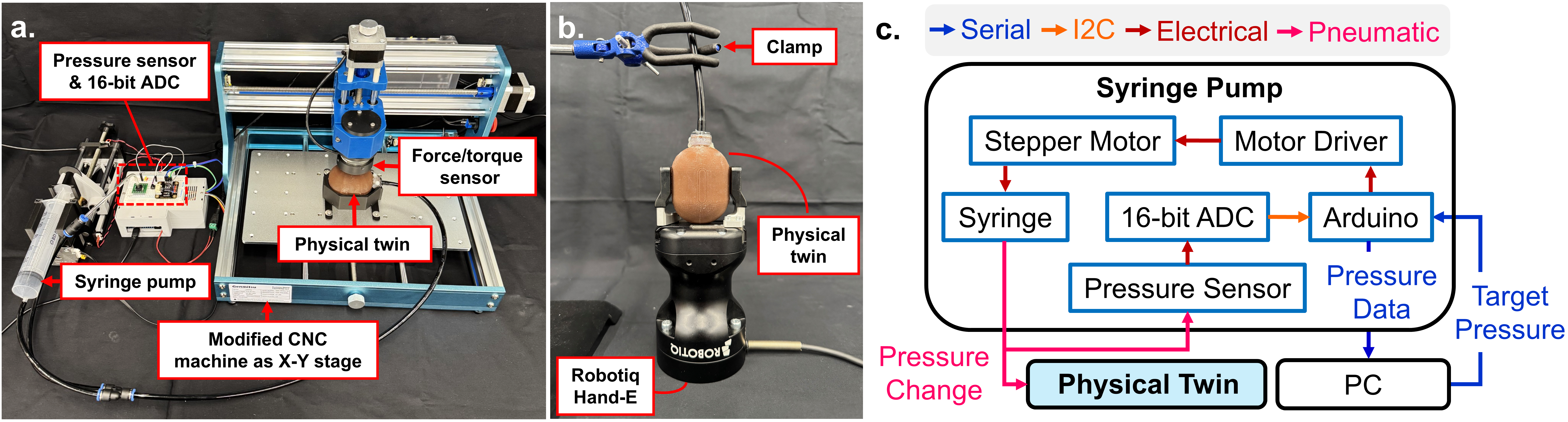}
    \caption{Hardware setup and system architecture of the pneumatic pressure control platform. (a) Experimental setup showing the physical twin mounted on a modified CNC machine serving as an X–Y stage, integrated with a force/torque sensor, a syringe pump, a pressure sensor, and a 16-bit ADC for data acquisition. (b) Gripping experiment using the Robotiq Hand-E gripper with the developed physical twin secured in a clamp. (c) System schematic of the physical twin illustrating electrical, I²C, serial, and pneumatic interconnections between the syringe pump, sensors, Arduino, and PC used to pressurize the physical twin for stiffness tuning.}
    \label{Exp_setup_fig}
\end{figure*}

\begin{figure*}[h!]
    \centering
    \includegraphics[scale=0.64]{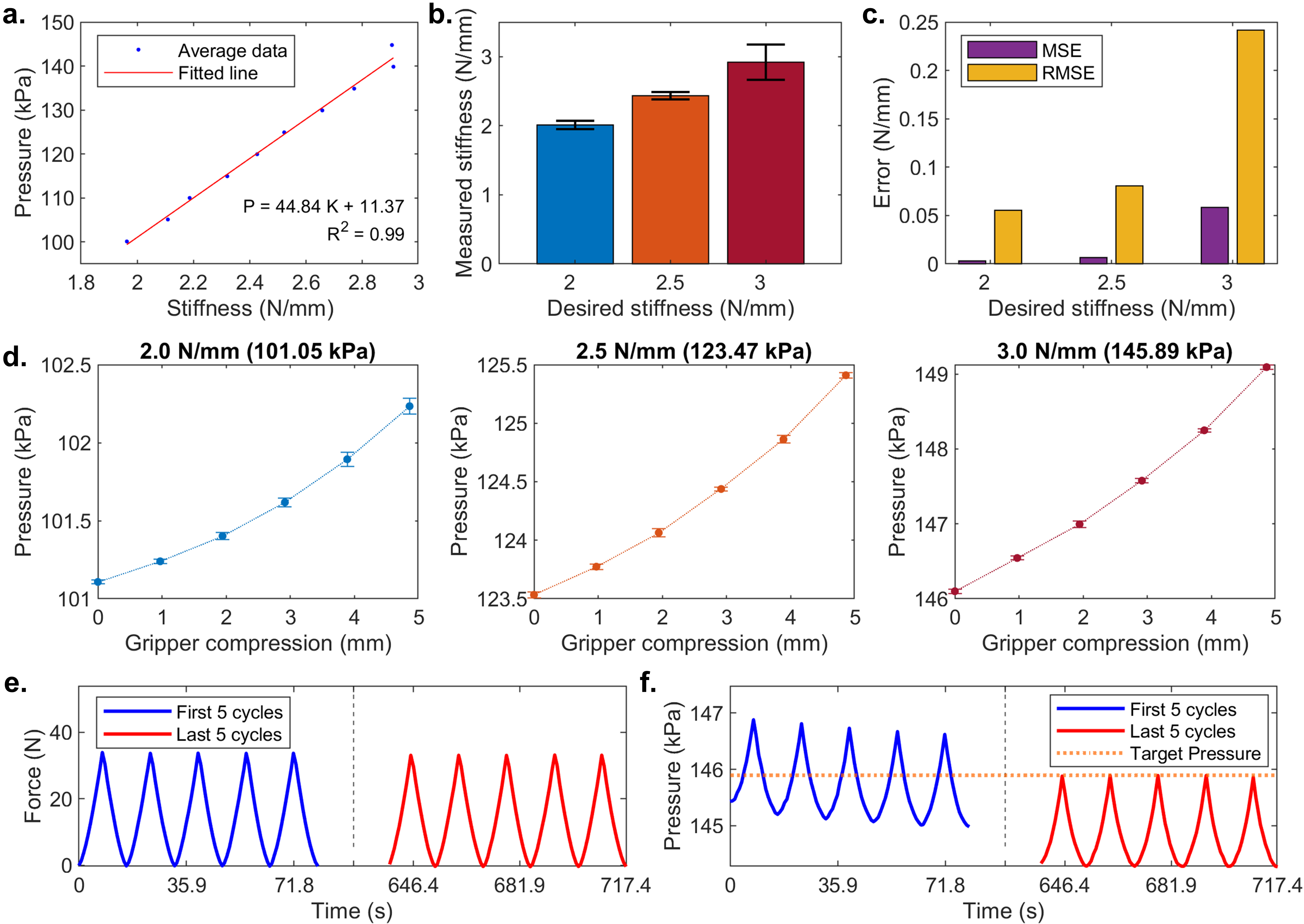}
    \caption{(a) The average pressure data from 5 trials (blue dots) against the measured stiffness is shown here. The red line shows the fitted first-order polynomial. The pressure data standard deviations were negligible. (b) The average measured stiffness across 5 trials is shown here, compared with the desired stiffness the physical twin was tuned to achieve. The black bars represent the standard deviations in the measured stiffness. (c) The MSE and RMSE between the measured and desired stiffnesses are shown here. (d) The average pressure values measured from the physical twin at each gripper compression step are shown here, based on the established stiffness level. The error bars on each point represent the standard deviation in the pressure readings. (e) The force measured with the force/torque sensor during the first and last 5 loading and unloading cycles (refer to legend) is shown here. (f) The internal pressure of the physical twin during the first and last 5 loading and unloading cycles is shown here.}
    \label{Results_fig}
\end{figure*}

\section{EXPERIMENTS and RESULTS} \label{experiments_sec}

\subsection{Stiffness tuning}\label{stiffness_tuning_exp}

To tune the physical twin to a desired stiffness, a relationship between the input pressure and the resulting stiffness must be determined. To achieve this, we used a syringe pump to pressurize the physical twin. A modified CNC router (Genmitsu 3018-PROVer V2, SainSmart, Inc.) with a force/torque sensor (RFT40, Robotous, Inc.) attached to its tool head was used as an X-Y stage to perform indentation experiments. This setup allows for intuitive stiffness measurements while staying synchronized with other hardware. First, the physical twin was pressurized to 100 kPa to maintain consistency throughout all trials. The pressure readings were acquired from a pressure sensor (MPXHZ6400A, 100-400 kPa, Freescale Semiconductors, Inc.). Then, a pre-load of 0.2 N was applied using the CNC router. The physical twin was indented by 3 mm using a flat-tip indenter attached to the force/torque sensor. Then, the force and pressure values were recorded. After that, the tool head of the CNC router was retracted to its starting position. The pressure was increased by 5 kPa, and the sequence was continued until a pressure of 145 kPa was reached. We conducted 5 trials in total, and the entire system was reset after each trial. The experiment setup is shown in Fig. \ref{Exp_setup_fig}(a). The syringe pump used to pressurize the physical twin consisted of a NEMA-17 stepper motor driven by a DRV8825 driver, which was controlled through an Arduino Uno R4. The voltage signals from the pressure sensor were captured with a 16-bit Analog-to-Digital Converter (ADC) (ADS1115, DFRobot, Inc.). The Arduino communicated with the ADC via the I²C interface. The entire system was synchronized and controlled by a computer executing a MATLAB 2024a (MathWorks, Inc.) script (refer to Fig. \ref{Exp_setup_fig}(c) for system schematic).

The z-axis force values (most significant) at 3 mm indentation were used to find the stiffness $(K = F_{R}/d)$ of the physical twin. The mean stiffness and pressure values acquired from the experiment were used to develop the calibration model. The MATLAB Curve Fitter app was used to determine the appropriate analytical model. A first-order polynomial model $(P = 44.84 K + 11.37)$, showed a good approximation $(R^2 = 0.9916)$ between pressure $(P)$, and stiffness. The calibration data is shown in Fig. \ref{Results_fig}(a). The standard deviations in the pressure readings were negligible.

This experiment showed that the physical twin reached a maximum stiffness of 3 N/mm at a pressure of 145 kPa. This stiffness was lower compared to that of the real Kiwi fruits when they were ripe (Day 9). This limitation is due to several factors, including the material's stiffness, the maximum achievable pressure, and the use of fiber reinforcement. Although this may be a limitation of the current physical twin, the objective of this initial study to develop a stiffness-tunable physical twin remains unchanged. With this limitation in mind, we chose 2, 2.5, and 3 N/mm as the desired stiffness values. These values simulate ripe, mid-ripe, and raw stages of a Kiwi. Using the calibration model, the required pressure levels to achieve the desired stiffnesses were 101.05, 123.47, and 145.89 kPa. To evaluate the model's accuracy, we performed indentation up to 3 mm when the physical twin was pressurized to achieve the desired stiffness levels. We performed 5 trials per desired stiffness level. The average measured stiffnesses of the physical twin (determined with force from the force/torque sensor) were compared with the desired stiffnesses in Fig. \ref{Results_fig}(b). The physical twin achieved stiffness accuracies of 99.43, 97.40, and 97.35\% when tuned to target stiffness values of 2, 2.5, and 3 N/mm, respectively. The Mean Squared Error (MSE) and Root Mean Squared Error (RMSE) between the desired and measured stiffness are shown in Fig. \ref{Results_fig}(c). This demonstrates that the calibration method is robust and accurate in tuning the physical twin to achieve the desired stiffness level.

\subsection{Sensor feedback with robotic gripping} \label{Gripper_exp_sec}

We used a Robotiq Hand-E gripper to perform stepwise grasping of the physical twin at stiffness levels of 2, 2.5, and 3 N/mm. The gripper was used to perform each grasp in 1 mm steps up to 5 mm at each stiffness level. First, to achieve a firm grip on the physical twin, the gripper was closed by 4 mm after fully opening. The pressure reading from the physical twin and the gripper compression were recorded at each step. We conducted 5 trials per desired stiffness level, and a Python script was used for synchronization and data acquisition. The experiment setup is shown in Fig. \ref{Exp_setup_fig}(b). The average pressure measured at each stiffness setting against the gripper compression is shown in Fig. \ref{Results_fig}(d). The low standard deviations indicate that the pressure readings obtained across multiple trials are consistent. It is clear that the pressure values increase as the gripper closes. This is expected as the gripper will apply more force to achieve the target compression. This demonstrates that internal sensor feedback from the physical twin can be used to determine gripping forces and understand how it is being handled.

\subsection{Stress test}

To evaluate the repeatability of the physical twin and its stiffness variation under high loads, we performed a stress test using the modified CNC machine setup. This experiment provides insight into whether the physical twin is suitable for repetitive grasping and can withstand high-force, faulty grasps. The physical twin was pressurized to the maximum stiffest setting we achieve (3 N/mm, 145.89 kPa). A pre-load of 0.2 N was applied to the physical twin. For this experiment, we mounted a 3D-printed flat plate with a diameter of 49mm into the force/torque sensor. This is to apply force over a greater area on the physical twin. Stepwise compression was performed to make data synchronization more intuitive. The physical twin was compressed in steps of 0.5 mm up to 5 mm (loading). Then, the tool head of the CNC router was retracted in 0.5 mm steps until the start position (unloading). This is considered a cycle, and we performed 50 compression cycles in total. The force data from the force/torque sensor and the pressure data from the physical twin were recorded at each step.

The first and last five cycles of the stress test are discussed in this section for intuitive interpretation. The measured force is shown in Fig. \ref{Results_fig}(e), and the internal pressure of the physical twin is shown in Fig. \ref{Results_fig}(f). To evaluate the difference between the first and last five cycles, the \textit{findpeaks} function in MATLAB was used to find their local maxima. The averages of the local maxima from the first and last five cycles were used to find the differences. The measured force showed a difference of only 0.64 N, and the internal pressure showed a difference of 0.86 kPa. The average local minima of the first and last cycles are 0.0056 kPa and 0.110 kPa, respectively. Therefore, the absolute percentage errors computed from the averages of the local minima of the target pressure (145.89 kPa) were 0.56\% and 1.10\% when the physical twin was fully uncompressed. The physical twin is repeatable under large loads (the average maximum applied force during the first 5 cycles is 33.74 N) during continuous, prolonged testing. This means the physical twin can maintain its stiffness consistently for grippers/manipulators that leverage data-driven approaches and require rigorous benchmarking. These methods require large datasets comprising data from multiple grasps to optimize grasping performance. The physical twin could improve the efficiency of the training process by reducing food waste, financial costs, and the effort required to use real fruits. 

\section{LIMITATIONS AND FUTURE WORK}\label{limitations_future_work_sec}

A limitation of our current work was the inability to achieve the stiffness levels of real Kiwi fruits. However, factors such as material stiffness, thickness, pneumatic actuation methods, and reinforcement techniques can all affect the maximum stiffness the physical twin can achieve. The sheer permutations of these factors can be difficult to explore in a single study. In the future, our objective is to explore improved manufacturing techniques and actuation methods to achieve stiffness values identical to those of characterized fruits. Furthermore, sensorizing physical twins using methods such as vision-based deformation tracking can enable processing of physical interactions, provide intrinsic tactile information, and reduce calibration complexity. In future work, we will explore the proposed methodology with fruits such as strawberries and bananas, which are more geometrically complex than a Kiwi. Additionally, we will explore the applicability of physical twins for more complex training tasks, such as slip detection and texture classification. We also aim to extend this concept to medical applications, such as replicating soft tissue properties with physical twins for palpation training.

\section{CONCLUSIONS}\label{Conclusion_sec}

 In this paper, we presented an initial prototype of a tunable physical twin that can adjust its stiffness. We characterized the stiffness of real Kiwi fruits, which guided our design and fabrication process. The physical twin was calibrated to achieve a pressure level corresponding to the desired stiffness. The calibration method was evaluated with three desired stiffness levels that mimic different levels of ripeness. Finally, we performed a gripping task with a commercially available gripper at different stiffness levels. This experiment showed that sensor feedback from the physical twin can be used to determine gripping forces. This can enable real-time adaptation of the gripping forces and prevent damage to the fruits. Furthermore, we performed a stress test on the physical twin, which demonstrated robust maintenance of the desired stiffness across multiple loading and unloading cycles. This experiment showed that the physical twin could withstand high gripping forces from grippers/manipulators during benchmarking and the acquisition of training data for data-driven control policies, thereby fostering sustainable testing.

\bibliographystyle{IEEEtran}  
\bibliography{IEEEexample}  
\end{document}